\pdfoutput=1
\documentclass[11pt]{article}
\usepackage[numbers,super,sort&compress]{natbib}
\usepackage[utf8]{inputenc}
\usepackage[T1]{fontenc}
\usepackage{lmodern}
\usepackage{amsmath,amssymb}
\usepackage{graphicx}
\usepackage{booktabs,tabularx,longtable}
\usepackage{siunitx}
\usepackage[hidelinks]{hyperref}
\usepackage[capitalise]{cleveref}
\usepackage{xcolor}
\usepackage[margin=2.5cm]{geometry}
\usepackage{caption}
\usepackage{enumitem}
\usepackage{microtype}
\usepackage{newfloat}
\crefname{figure}{Fig.}{Figs}
\Crefname{figure}{Fig.}{Figs}
\crefname{section}{Section}{Sections}
\newcommand{\ci}[2]{[#1,\,#2]}
\DeclareUnicodeCharacter{2212}{\ensuremath{-}}
\DeclareUnicodeCharacter{00D7}{\ensuremath{\times}}
\DeclareUnicodeCharacter{2192}{\ensuremath{\rightarrow}}
\DeclareUnicodeCharacter{2190}{\ensuremath{\leftarrow}}
\DeclareUnicodeCharacter{2265}{\ensuremath{\geq}}
\DeclareUnicodeCharacter{2264}{\ensuremath{\leq}}
\DeclareUnicodeCharacter{2248}{\ensuremath{\approx}}
\DeclareUnicodeCharacter{00B1}{\ensuremath{\pm}}
\DeclareUnicodeCharacter{03C1}{\ensuremath{\rho}}
\DeclareUnicodeCharacter{0394}{\ensuremath{\Delta}}
\DeclareUnicodeCharacter{03BA}{\ensuremath{\kappa}}
\DeclareUnicodeCharacter{03B3}{\ensuremath{\gamma}}
\DeclareUnicodeCharacter{03BE}{\ensuremath{\xi}}
\DeclareUnicodeCharacter{2032}{\ensuremath{'}}
\DeclareUnicodeCharacter{2009}{\,}
\DeclareUnicodeCharacter{202F}{\,}

\usepackage{float}

\crefname{section}{Section}{Sections}
\Crefname{section}{Section}{Sections}
\crefname{subsection}{Section}{Sections}
\Crefname{subsection}{Section}{Sections}
\crefname{subsubsection}{Section}{Sections}
\Crefname{subsubsection}{Section}{Sections}
\makeatletter\@ifundefined{NC@find@Y}{\newcolumntype{Y}{>{\raggedright\arraybackslash}X}}{}\makeatother
\makeatletter
\renewcommand*\l@section[2]{\ifnum\c@tocdepth>\m@ne\addpenalty{-\@highpenalty}\vskip 1.0em\@plus\p@
  \setlength\@tempdima{3.0em}\begingroup\parindent\z@\rightskip\@pnumwidth\parfillskip-\@pnumwidth
  \leavevmode\bfseries\advance\leftskip\@tempdima\hskip-\leftskip#1\nobreak\hfil\nobreak\hb@xt@\@pnumwidth{\hss#2}\par\penalty\@highpenalty\endgroup\fi}
\renewcommand*\l@subsection{\@dottedtocline{2}{3.0em}{3.8em}}
\renewcommand*\l@subsubsection{\@dottedtocline{3}{6.8em}{4.8em}}
\makeatother

\providecommand{\snTabStart}{\par\medskip\noindent\begin{minipage}{\textwidth}\centering}
\providecommand{\snTabEnd}{\end{minipage}\par\medskip}

\crefname{figure}{Fig.}{Figs}
\Crefname{figure}{Fig.}{Figs}
\crefname{table}{Table}{Tables}
\Crefname{table}{Table}{Tables}
\crefname{section}{Section}{Sections}

\renewcommand{\figurename}{Fig.}
\DeclareCaptionLabelSeparator{nsbar}{ $|$ }
\newsavebox{\arxivlegbox}
\newlength{\arxivimght}
\newcommand{\arxivfloatfig}[5]{%
\begin{#1}[p]\centering
\sbox{\arxivlegbox}{\begin{minipage}{\textwidth}\caption[#3]{\textbf{#3} #5}\label{#4}\end{minipage}}%
\setlength{\arxivimght}{\dimexpr\textheight-\ht\arxivlegbox-\dp\arxivlegbox-2\baselineskip\relax}%
\includegraphics[width=\textwidth,height=\arxivimght,keepaspectratio]{#2}\par\vspace{6pt}
\usebox{\arxivlegbox}
\end{#1}}
\usepackage{afterpage}
\makeatletter
\newtoks\arxq@toks
\newtoks\arxq@ttoks
\newif\ifarxq@armed
\let\arxq@item\relax
\newwrite\arxq@log
\immediate\openout\arxq@log=\jobname.figpos
\long\def\arxivqueue#1{\global\arxq@toks\expandafter{\the\arxq@toks\arxq@item{#1}}\arxq@arm}
\long\def\arxivqueuetab#1{\arxivcitemark{table}\global\arxq@ttoks\expandafter{\the\arxq@ttoks\arxq@item{#1}}\arxq@arm}
\def\arxq@arm{\ifarxq@armed\else\global\arxq@armedtrue\afterpage{\arxq@pop}\fi}
\def\arxq@pop{\edef\@tempa{\the\arxq@toks}%
  \ifx\@tempa\@empty\expandafter\@firstoftwo\else\expandafter\@secondoftwo\fi{\arxq@popt}{\arxq@popf}}
\def\arxq@popf{\expandafter\arxq@split\the\arxq@toks\arxq@end}
\def\arxq@popt{\expandafter\arxq@splitt\the\arxq@ttoks\arxq@end}
\long\def\arxq@split\arxq@item#1#2\arxq@end{\global\arxq@toks{#2}#1\par\arxq@tabafter\arxq@next}
\def\arxq@tabafter{\edef\@tempa{\the\arxq@ttoks}%
  \ifx\@tempa\@empty\else\expandafter\arxq@splitti\the\arxq@ttoks\arxq@end\fi}
\long\def\arxq@splitti\arxq@item#1#2\arxq@end{\global\arxq@ttoks{#2}%
  \write\arxq@log{table start \thepage}#1\par\write\arxq@log{table end \thepage}\vspace{\baselineskip}}
\long\def\arxq@splitt\arxq@item#1#2\arxq@end{\global\arxq@ttoks{#2}%
  \write\arxq@log{table start \thepage}#1\par\write\arxq@log{table end \thepage}\arxq@next}
\def\arxq@next{\edef\@tempa{\the\arxq@toks\the\arxq@ttoks}%
  \ifx\@tempa\@empty\global\arxq@armedfalse\else\afterpage{\arxq@pop}\fi}
\def\arxivcitemark#1{\write\arxq@log{#1 cited \thepage}}
\newlength{\arxivmainimght}
\long\def\arxivmainfig#1#2#3#4{%
  \par\begingroup\centering\refstepcounter{figure}\label{#3}\write\arxq@log{#3 start \thepage}%
  \setlength{\arxivmainimght}{\dimexpr\textheight-2\baselineskip\relax}%
  \includegraphics[width=\textwidth,height=\arxivmainimght,keepaspectratio]{#1}\par\endgroup
  \vspace{6pt}%
  \begingroup\footnotesize\setlength{\parindent}{0pt}\setlength{\parskip}{0pt}%
  \noindent\textbf{\figurename~\thefigure\ $|$ }\textbf{#2} #4\par\endgroup
  \write\arxq@log{#3 end \thepage}%
  \vspace{\baselineskip}}
\makeatother
\makeatletter
\newcommand{\arxivfig}[5]{\def\@tempa{#1}\def\@tempb{figure}%
  \ifx\@tempa\@tempb\expandafter\@firstoftwo\else\expandafter\@secondoftwo\fi
  {\arxivcitemark{#4}\arxivqueue{\arxivmainfig{#2}{#3}{#4}{#5}}}{\arxivfloatfig{#1}{#2}{#3}{#4}{#5}}}
\IfFileExists{latexml.sty}{\usepackage{latexml}}{}
\@ifundefined{iflatexml}{\newif\iflatexml\latexmlfalse}{}
\iflatexml
  \renewcommand{\arxivfig}[5]{\begin{#1}[t]\centering
    \includegraphics[width=\textwidth]{#2}\caption[#3]{\textbf{#3} #5}\label{#4}\end{#1}}
  \long\def\arxivqueuetab#1{#1}
  \def\arxivcitemark#1{}
  \newcommand{\arxivtoc}[2]{\addcontentsline{toc}{#1}{#2}}
\else
  \newcommand{\arxivtoc}[2]{}
\fi
\makeatother
\makeatletter
\providecommand{\FloatBarrier}{\par\begingroup\let\@elt\relax
  \edef\@tempa{\@botlist\@deferlist\@dbldeferlist}\def\@tempb{}%
  \ifx\@tempa\@tempb\endgroup\else\endgroup\clearpage\fi}
\makeatother

\hypersetup{pdftitle={A research world model of power engineering learns its dynamics and reveals what-ifs},
  pdfauthor={Fanping Sui, Ying Zhang, Liwei Lin}}

\title{A research world model of power engineering learns its dynamics and reveals what-ifs}
\author{Fanping Sui\textsuperscript{1}, Ying Zhang\textsuperscript{2,*} and Liwei Lin\textsuperscript{1}\\[6pt]
\small\textsuperscript{1}Department of Mechanical Engineering, University of California, Berkeley, Berkeley, CA, USA\\
\small\textsuperscript{2}School of Electrical and Computer Engineering, Oklahoma State University, Stillwater, OK, USA\\
\small\textsuperscript{*}Corresponding author. Email: \href{mailto:y.zhang@okstate.edu}{y.zhang@okstate.edu}}
\date{}
\hypersetup{pdftitle={A research world model of power engineering learns its dynamics and reveals what-ifs},
  pdfauthor={Fanping Sui, Ying Zhang, Liwei Lin}}

\begin{document}
\maketitle

\begin{abstract}
\noindent
Decisions about the direction of research in power engineering determine which technologies mature and, through them, shape infrastructure that lasts decades, yet their consequences cannot be observed before they are taken, and existing forecasts of research fields extrapolate a single trajectory. Here we introduce research world models, which treat a research field as a dynamical system that can be acted upon, and establish one for power engineering from 1.1 million papers. A distilled language model reads each paper into role-typed concepts, two sophisticated, high-fidelity graph--temporal neural network cores learn how the resulting concept network evolves, and operators learned from natural experiments in the publication record estimate the effects of interventions. In temporally held-out tests, the model forecast years ahead which concepts would grow and which method--target pairs would be joined for the first time, outperforming tuned machine-learning baselines in most settings, and its estimated intervention effects tracked those realized in natural experiments it had not seen. Its training scheme and three of its operators replicated in two fields withheld from development. Questions posed in natural language are compiled by a domain-specialized 27-billion-parameter language model into executable actions and answered as paired simulations with and without the intervention. The answers were rated 4.1 of 5 for faithfulness by an independent language-model judge. By revealing the latent dynamics of a research field and its response to interventions, research world models offer a new way to simulate and test research decisions before they are made.
\end{abstract}

\section*{Introduction}\label{sec:intro}\arxivtoc{section}{Introduction}

Power engineering research supports infrastructure that is long-lived, capital-intensive and coupled to nearly every sector of the economy. The technologies and methods it develops today therefore shape the options of energy systems for decades\cite{iea2021netzero}. Because such systems are strongly path-dependent, with early technological choices becoming locked in through investment and institutions\cite{arthur1989competing,unruh2000understanding}, misallocated research effort is costly and difficult to reverse. Even the technologies now central to the energy transition illustrate these timescales: photovoltaics and lithium-ion batteries required several decades of research and development to progress from laboratory demonstration to deployment at scale, and their costs declined only as cumulative production grew\cite{ziegler2021reexamining,way2022empirically}. The research itself evolves through internal dynamics, in which attention concentrates on established topics and new problem--method combinations emerge only gradually\cite{foster2015tradition,uzzi2013atypical}, and it is repeatedly redirected by general-purpose technologies, from electrification to semiconductors and, most recently, machine learning, each of which changed both the problems the field addresses and the methods it uses\cite{bresnahan1995general,cockburn2019impact}. Shifts in policy priorities and funding act on the same trajectory\cite{johnstone2010renewable,myers2020elasticity}. Deciding on an intervention in such a field therefore requires both a forecast of its course and an estimate of how that course would change under the intervention. Yet this counterfactual is inaccessible to direct observation: each intervention is enacted only once, and the trajectory it forecloses is never realized, an instance of the fundamental problem of causal inference\cite{holland1986statistics}. The instruments currently available, expert elicitation and bibliometric indicators, are retrospective by construction and silent on such counterfactuals\cite{verdolini2018future,fortunato2018science}.

In machine learning, the problem of anticipating how a system evolves under alternative actions is the province of world models. A world model infers a latent state from incomplete observations and learns its transition
\begin{equation}
S_{t+1} = T(S_t, a_t),
\label{eq:transition}
\end{equation}
where $S_t$ is the latent state at step $t$ and $a_t$ an action\cite{ha2018worldmodels,hafner2025mastering}. Because forecasting and counterfactual evaluation draw on a single learned transition, the two questions are answered within one model. Learned simulators of this kind now outperform operational numerical systems in medium-range weather prediction\cite{lam2023graphcast,price2025gencast} and have been extended to the Earth system\cite{bodnar2025aurora} and to human physiology\cite{wang2026world}.

A research field has a latent state of this kind: the problems it pursues and the methods brought to bear on them are never recorded as such, but are reflected, incompletely and with delay, in its publications. The science of science has forecast topic growth and the emergence of links between concepts\cite{krenn2023forecasting,marwitz2026predicting}, and dynamic knowledge-graph pipelines have been proposed for the same purpose\cite{arman2026dynscigraph}, yet these approaches extrapolate a single trajectory: they do not learn a transition of the field under interventions and have not been validated against the observed effects of interventions. Such validation is difficult because concept growth is dominated by persistence and mean reversion, which makes naive baselines formidable, and because interventions are unlabeled, so that their effects are identifiable only from natural experiments\cite{imbens2015causal}. Artifacts of the record, such as duplicated entries, can also be mistaken for genuine dynamics.

Here we introduce the research world model, a world model in which the state $S_t$ of equation~(\ref{eq:transition}) is a research field, represented by its typed concept hypergraph and the latent embedding of that hypergraph, the transition $T$ is learned from the publication record, the actions $a_t$ are research interventions, such as a burst of papers, a new term or the adoption of a method, and the answer to a what-if question is a pair of simulations with and without the action. To our knowledge, no earlier model of a research field has learned its transition under interventions. We build a research world model for power engineering, with semiconductors and materials science as replication fields. We test whether its latent state forecasts concept growth and new method--target links more accurately than tuned baselines, whether its training scheme and operators transfer to fields not used in development, and whether its operators recover held-out natural experiments. The experiments also establish how far each answer is identified: bursts of papers on first-time method--target pairs have identified effects, and releases, retractions and policy events have partly identified effects. We further test whether natural-language questions can be answered without a language model performing any computation. Duplicate records (4.5\,\% of records but 83\,\% of naively detected bursts) were removed from the forecasting evaluations and from the detection of bursts. Each evaluation followed a pre-registered decision rule\cite{nosek2018preregistration}, and comparisons of trained components used at least five independently trained runs per arm with a run-level bootstrap. A forecast made from December 2025 was frozen for later scoring against observed data.

\section*{Results}\label{sec:results}\arxivtoc{section}{Results}

\subsection*{Latent state from the paper stream}\label{sec:res-architecture}\arxivtoc{subsection}{Latent state from the paper stream}

The research world model reads 1.20 million power-engineering abstracts (1.14 million papers after deduplication), the 11.1 million citation events of the citation layer and the use of their terms in papers from seven other fields (\cref{fig:system}a). A distilled language-model reader extracts from each abstract its role-typed concepts and their relations, which form a monthly typed hypergraph of 204,287 concepts with one hyperedge per deduplicated paper. This hypergraph is read by two temporal graph networks, the growth core and the link core. The hypergraph and the concept states that these networks compute from it form the state $S_t$ of equation~(\ref{eq:transition}). The growth core forecasts the log growth of each concept 12, 24, 36 and 60 months ahead (\cref{fig:system}a,b,c(i)). On the field map, four links between regions that the link core ranked at December 2019 (ranks 1,551--3,996) were first named in a title or abstract in 2020--2023 (\cref{fig:system}b). On the states of the link core, a link readout forecasts which concepts will be linked for the first time (\cref{fig:system}c(ii)), for all 15 pairs of the five concept roles (problem, method, object, setting and metric).

\arxivfig{figure}{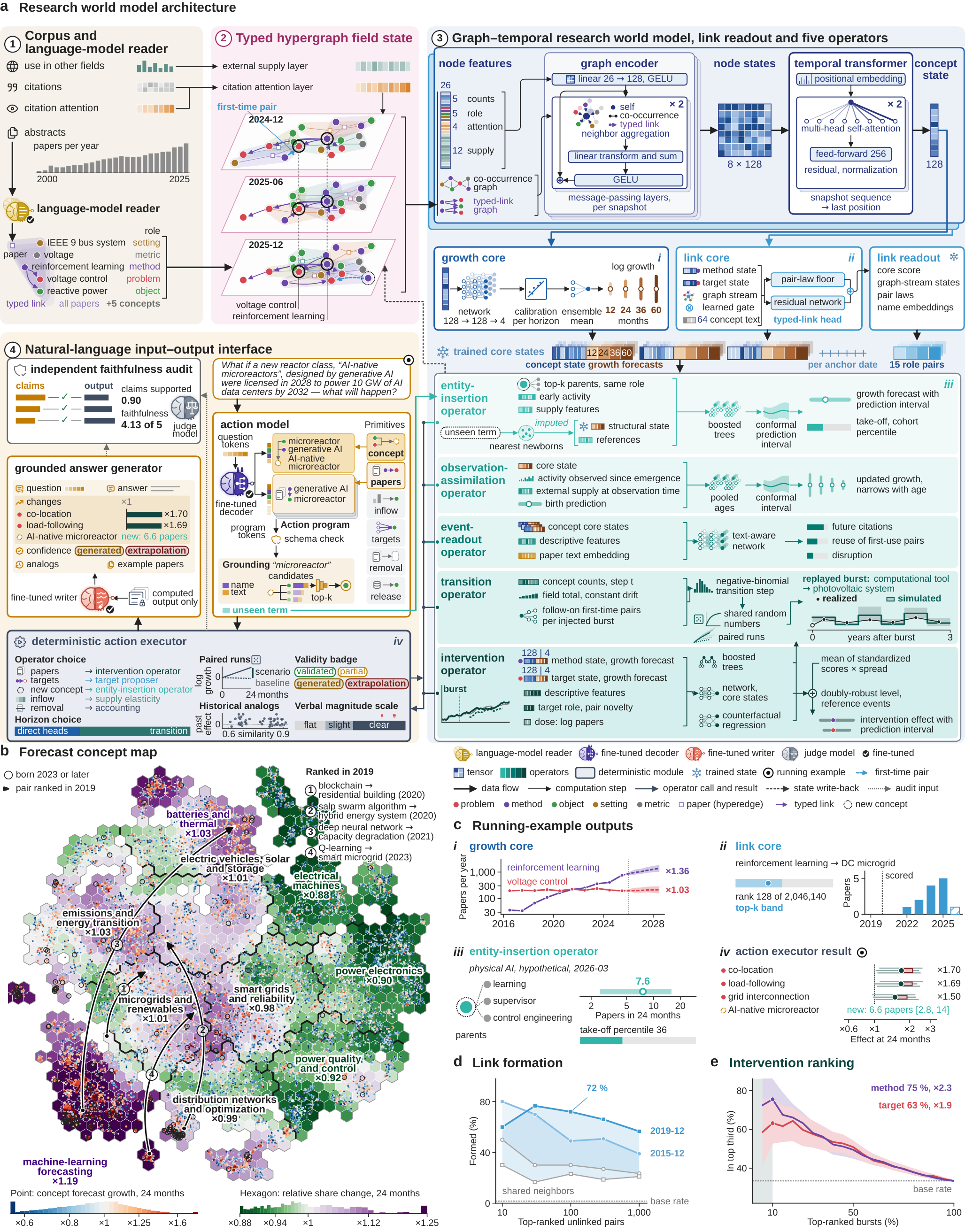}{A research world model of power engineering.}{fig:system}{%
\textbf{a},~Architecture (schematic, with tensor dimensions as implemented). (1)~A 7-billion-parameter reader distilled from a 27-billion-parameter teacher converts 1.20 million abstracts into role-typed concepts. Citations and use in seven other fields form the citation-attention and external-supply layers. (2)~The field state, a monthly typed hypergraph (real subgraph around the running example). (3)~The growth and link cores (stacked cards) share one architecture, in which a graph encoder and a temporal transformer return a 128-dimensional concept state, and are trained separately. The link core adds a graph stream with a learned gate, concept text and a typed-link head. One link readout on the link-core states scores pairs of all 15 role combinations, and five operators (teal shades) read the trained core states of both cores. (4)~Interfaces, shown for a what-if question (box, condensed). In the action model, a fine-tuned decoder compiles the question into an action program over state primitives, which passes a schema check and whose concepts are grounded. The deterministic action executor runs scenario and baseline on shared random numbers and returns historical analogs and a validity badge. The grounded answer generator (a fine-tuned writer) produces the answer from the computed output alone, and the independent judge model audits each claim.
\textbf{b},~Forecast concept map (t-SNE\cite{vandermaaten2008tsne}, December 2025) colored by 24-month forecast. Hexagons, forecast change in share of mentions. Circles, concepts born in 2023 or later. Arrows, four links between regions ranked at December 2019 before any title or abstract named both terms (ranks 1,551, 1,650, 3,239 and 3,996; first named 2020--2023).
\textbf{c},~Running example. c(i), growth forecasts of the growth core (80\,\% intervals). c(ii), a pair ranked 128 of 2,046,140 by the link core at December 2019 that formed later. c(iii), a hypothetical new term. c(iv), executor result for the question of a: 24-month effects on three concepts (lines, 80\,\% intervals of the three injected pairs per concept; box, level band; point, combined effect) and the entity-insertion forecast for the new concept (papers in 24 months, 80\,\% interval).
\textbf{d},~Top-ranked unlinked pairs formed within 24 months, against shared-neighbor ranking and the base rate.
\textbf{e},~Intervention operator on 574 test bursts: share of top-ranked bursts with a top-third realized effect (top tenth: method 75\,\%, 2.3 times the base rate of 33.4\,\%; target 63\,\%, $\times$1.9).}

Five operators read the trained core states and answer what-if questions according to their subject. The entity-insertion operator adds a term that does not yet exist and forecasts its trajectory (\cref{fig:system}c(iii)), and the observation-assimilation operator revises that forecast once the first papers using the term have appeared. The event-readout operator predicts the citations, pair reuse and disruption\cite{funk2017dynamic} of a single paper. The intervention operator estimates how a burst of injected activity changes concept states, and the estimators for the other what-if events are its variants. The transition operator advances the whole field as paired scenario and baseline trajectories.

A question in natural language reaches the operators through a fine-tuned decoder, which compiles it into an action program, either a registered action of one of 18 types or a composition of six state primitives, and grounds every action to existing concepts. The program is run by the deterministic action executor once as scenario and once as baseline with shared random numbers. It returns their difference with intervals, historical analogs and a validity badge that states how far the effect is identified (``validated'', ``partial'', ``extrapolation'', ``not supported'' or, at level 2, ``generated''). A grounded answer generator (a fine-tuned writer) produces the answer from this output alone (labeled ``computed output'' in \cref{fig:system}a), and an independent judge model audits the answer claim by claim. The what-if question of \cref{fig:system}a lies outside the registry. The fine-tuned decoder compiles it, at level 2 of the action model, into an added concept and bursts of papers (Methods), and the executor returns 24-month effects of 1.50--1.70 times the baseline on three concepts and 6.6 papers for the new concept (\cref{fig:system}c(iv)). \Cref{fig:system}d,e summarizes the link-formation and intervention results reported below: the share of top-ranked unlinked pairs that formed within 24 months, against shared-neighbor ranking and the base rate, and the share of the top-ranked test bursts of the intervention operator whose realized effect fell in the top third.

\subsection*{Forecasting concept growth}\label{sec:res-growth}\arxivtoc{subsection}{Forecasting concept growth}

Growth forecasts from the learned state were compared with those of strong conventional regressors at five walk-forward cutoffs (2013--2021) and four horizons, in 19 cells with complete labels. Each growth head is trained with a horizon-specific cutoff and learns from every past month (anchor) whose outcome at its own horizon had been observed by the cutoff (\cref{fig:forecasting}a and Methods).

\arxivfig{figure}{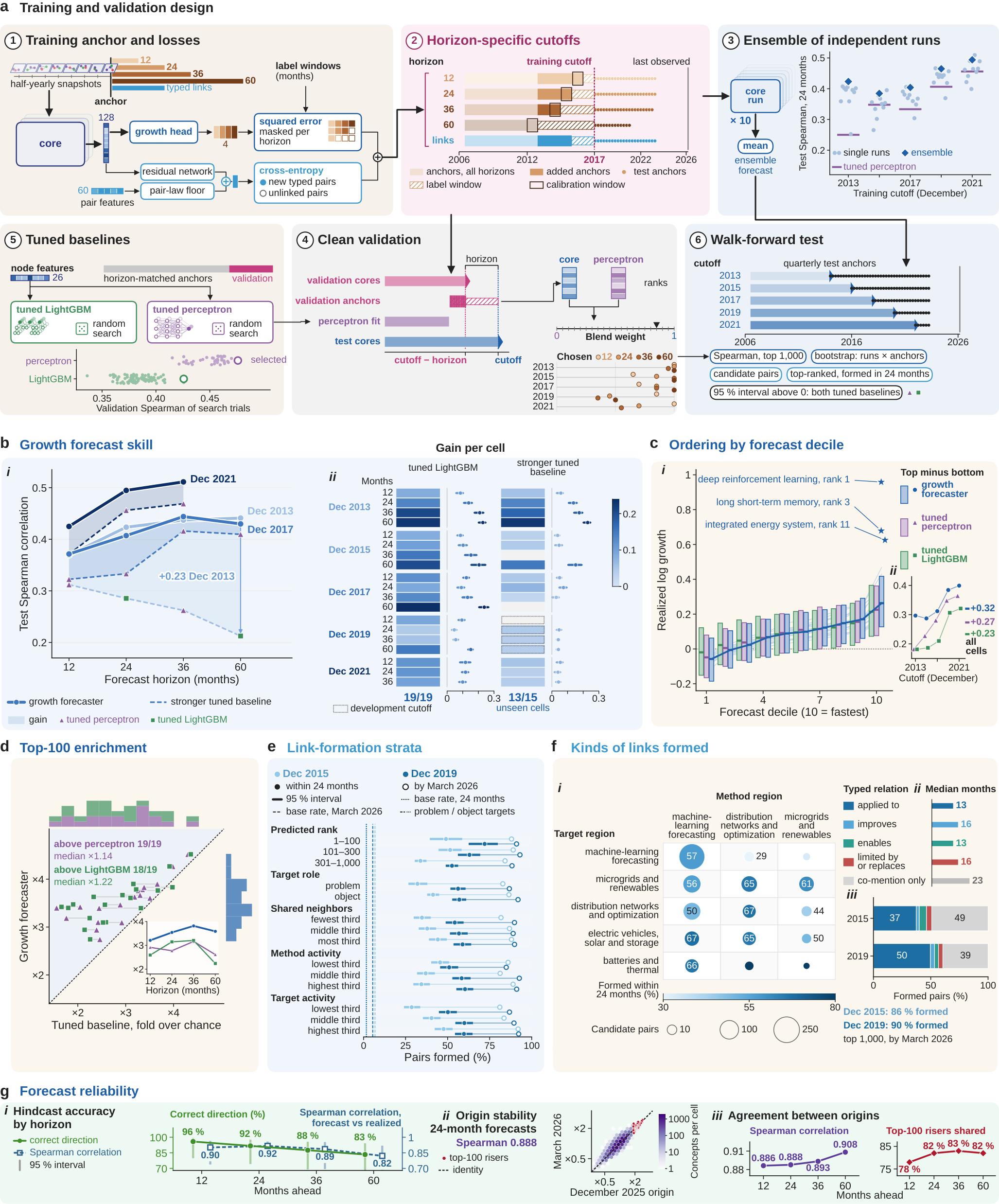}{Training, validation and test of growth and link forecasting.}{fig:forecasting}{%
\textbf{a},~Design, with mini-plots of results. (1)~A training anchor with masked growth losses and a typed-link loss. (2)~Horizon-specific training cutoffs. (3)~Ten independent runs. (4)~Blend weight chosen on validation cores trained at the cutoff minus the horizon. (5)~Tuned baselines on the same 26 node features. (6)~Walk-forward test at five cutoffs.
\textbf{b},~b(i), test Spearman correlation over the 1,000 most active concepts by horizon for three cutoffs, growth forecaster (solid) against the stronger tuned baseline (dashed). b(ii), gain in each of 19 cells over tuned LightGBM and over the stronger tuned baseline (mean and 95\,\% interval where above 0). The forecaster exceeds tuned LightGBM in all 19 cells and both tuned baselines in 13 of the 15 unseen cells. December 2019 cells, seen during development, are shown for comparison.
\textbf{c},~c(i), realized log growth by forecast decile for the growth forecaster and the two tuned baselines on identical anchors. c(ii), top-minus-bottom decile median (+0.32, +0.27 and +0.23).
\textbf{d},~Top-100 enrichment over chance, growth forecaster against each tuned baseline per cell (37 of 38 points above the identity line).
\textbf{e},~Share of the top 1,000 unlinked pairs formed within 24 months (filled) and by March 2026 (open), by rank, target role, shared neighbors and activity.
\textbf{f},~f(i), December 2019 top 1,000 by map region of method and target. f(ii), months to formation per typed relation. f(iii), first relation extracted.
\textbf{g},~Reliability. g(i), hindcast direction (96, 92, 88, 83\,\%) and Spearman correlation (0.90, 0.92, 0.89, 0.82) by horizon. g(ii), origin stability of the five-run forecast (Spearman 0.888). g(iii), agreement between origins.}

The growth forecaster blends the ranking of ten horizon-specific runs with that of a tuned multilayer perceptron, and it is evaluated on the 15 cells of the cutoffs other than December 2019. In 13 of these cells it was better than both a multilayer perceptron and LightGBM\cite{ke2017lightgbm} tuned on the same node features, each time with a 95\,\% interval above zero (\cref{fig:forecasting}b). It also ranked growth better than a zero-shot language model. Its top-minus-bottom decile median of realized growth was +0.32, against +0.27 and +0.23 for the tuned perceptron and LightGBM (\cref{fig:forecasting}c), and its top-100 enrichment was higher in 37 of 38 comparisons (\cref{fig:forecasting}d). It exceeded default LightGBM at all 393 test anchors and, over all 19 cells, both tuned baselines in 16, with no interval entirely below zero.

\subsection*{Forecasting new links between concepts}\label{sec:res-links}\arxivtoc{subsection}{Forecasting new links between concepts}

Links between concepts depend on pair-level structure that node-feature regressors do not represent. The link core adds a gated graph stream, a typed-link head and concept text to represent it, and one link readout on its trained states scores pairs of every role combination. At all five cutoffs in power engineering, the link core had a higher typed average precision than the growth core, by +0.023 to +0.049, with every interval above zero. Growth and links are therefore forecast by separate cores, each trained for its own task (Methods). The link core is trained with horizon-specific cutoffs, under which its primary metric, hard-negative typed average precision, exceeds that of the default scheme by +0.023 \ci{+0.018}{+0.029}.

A retrospective test was run at December 2019 on every unlinked pair among the 1,500 most active methods and the 1,500 most active problems and objects, about 2.05 million pairs. Of the 100 pairs that the link core ranked highest, 72\,\% formed within 24 months, against a base rate of 1.75\,\% (\cref{tab:newlinks}). Over both cutoffs (December 2015 and 2019) and two seed sets, the top 100 formed at 35--41 times the base rate (\cref{fig:system}d). Under the strictest alias map, which also merges aliases matched by embeddings, the December 2019 share is 65\,\% instead of 72\,\%.

\arxivqueuetab{%
\begin{table}[H]
\caption{\textbf{Forecasting new method--target links}}\label{tab:newlinks}
\footnotesize
\setlength{\tabcolsep}{4pt}
\renewcommand{\arraystretch}{1.25}
\begin{tabular*}{\textwidth}{@{\extracolsep{\fill}}lllll@{}}
\toprule
 & \multicolumn{2}{c}{December 2015} & \multicolumn{2}{c}{December 2019} \\
\cmidrule(lr){2-3}\cmidrule(l){4-5}
 & within 24 months & by March 2026 & within 24 months & by March 2026 \\
\midrule
\multicolumn{5}{@{}l}{\textit{All unlinked pairs, precision at 100 (\%)}} \\
Link readout, horizon-specific link cores & 49 \ci{39}{59} & 88 \ci{82}{94} & 72 \ci{63}{81} & 91 \ci{85}{96} \\
Link readout, default-scheme link cores & 55 \ci{45}{64} & 91 \ci{85}{96} & 67 \ci{58}{76} & 87 \ci{80}{93} \\
Adamic--Adar & 30 \ci{22}{39} & 71 \ci{62}{79} & 23 \ci{15}{31} & 51 \ci{41}{61} \\
Base rate & 1.40 & 6.80 & 1.75 & 5.47 \\
Enrichment, horizon-specific link cores ($\times$) & 35.0 & 12.9 & 41.1 & 16.6 \\
\addlinespace
\multicolumn{5}{@{}l}{\textit{New pairs, precision at 100 (\%)}} \\
Link readout, horizon-specific link cores & 36 \ci{26}{45} & 89 \ci{82}{95} & 50 \ci{41}{59} & 92 \ci{86}{97} \\
Link readout, default-scheme link cores & 29 \ci{21}{38} & 83 \ci{75}{90} & 42 \ci{33}{52} & 85 \ci{78}{92} \\
Adamic--Adar & 18 \ci{11}{26} & 56 \ci{46}{66} & 15 \ci{8}{23} & 40 \ci{30}{50} \\
Base rate & 0.81 & 4.63 & 1.02 & 3.59 \\
Enrichment, horizon-specific link cores ($\times$) & 44.4 & 19.2 & 49.0 & 25.6 \\
\bottomrule
\end{tabular*}

\vspace{4pt}
\begin{minipage}{\textwidth}\footnotesize
Power engineering. Candidates, pairs of the 1,500 most active methods and the 1,500 most active problem or object concepts without a typed link at the cutoff (2,099,192 at December 2015 and 2,046,140 at December 2019). All unlinked pairs, all candidates. New pairs, candidates whose two concepts no earlier paper mentions together under any variant of their names and for which no earlier paper contains concepts whose names include both, ranked within this pool. Precision at 100, share of the 100 highest-ranked pairs that formed a typed method--target link under the exact concept names within 24 months of the cutoff or by March 2026. Link readout, scores of the readout on link cores trained with horizon-specific or with default-scheme training cutoffs (mean of five runs). Adamic--Adar\cite{adamic2003friends}, on the co-occurrence graph at the cutoff. Base rate, share of the candidates of the block that formed (\%). Enrichment, precision at 100 divided by the base rate. Square brackets, 95\,\% bootstrap intervals over the top 100 pairs.
\end{minipage}
\end{table}
}

A stricter test was run on new pairs, whose two concepts no earlier paper mentions together under any variant of their names and for which no earlier paper contains concepts whose names include both. Among the top 100 new pairs, the share that formed within 24 months was 36\,\% at December 2015 and 50\,\% at December 2019, and the share that formed by March 2026 was 89\,\% and 92\,\% (\cref{tab:newlinks}). Of the 1,000 highest-ranked unlinked pairs, 39\,\% (December 2015) and 57\,\% (December 2019) formed within 24 months (\cref{fig:forecasting}e,f(i)) and 86\,\% and 90\,\% by March 2026. At December 2019, formed pairs with an extracted relation took a median of 13--16 months, against 23 months for co-mention only (\cref{fig:forecasting}f(ii),(iii)).

First links between concepts of any two roles, for example between two methods, between a problem and a setting, or between an object and a metric, are scored by the same readout. In all 75 combinations of role pair and cutoff, the readout's average precision exceeded that of the strongest of 11 baselines, with an interval above zero in 74 (Methods).

In 24 hindcasts of concept growth from four earlier origins, the forecast direction was correct in 96, 92, 88 and 83\,\% of cases at 12, 24, 36 and 60 months (22 of 24 at 24 months; 18 hindcasts at 60 months; \cref{fig:forecasting}g(i)), and the five-run forecast agreed between origins at a Spearman correlation of 0.888 at 24 months (\cref{fig:forecasting}g(ii)). Between the two origins, the Spearman correlation was 0.886--0.908 at 12--60 months, and 78--83\,\% of the 100 highest-ranked risers were shared (\cref{fig:forecasting}g(iii)).

\subsection*{Operators on natural experiments}\label{sec:res-operators}\arxivtoc{subsection}{Operators on natural experiments}

Each operator was validated on natural experiments not used in training, against the strongest available baseline. The intervention operator learns from genuine bursts of papers on first-time method--target pairs, each matched to five control concepts in a difference-in-differences design\cite{card1994minimum}. On 186 blind test bursts, its predictions correlated with the realized effects at $r = 0.557$, which is 0.78 of the noise ceiling set by the spread of the matched controls (\cref{fig:operators}a). Its squared errors were smaller than those of injecting the burst into the core or of assuming a constant average effect. Across all 574 test bursts, Pearson correlations of the per-burst predictions were 0.48--0.56 (method) and 0.32--0.40 (target, over two seed sets), and the share with the predicted sign rose from 55\,\% to 87\,\% (method) and from 63\,\% to 80\,\% (target) between the lowest and the highest thirds of absolute predicted effect (\cref{fig:operators}a). On the method side, three-quarters of its top tenth had realized effects in the top third, 2.3 times the base rate (\cref{fig:system}e). Predicted means also tracked realized means within strata defined on training bursts, with correlations of 0.96 (2015) and 0.79 (2019) against a criterion of 0.6 (\cref{fig:operators}b).

\arxivfig{figure}{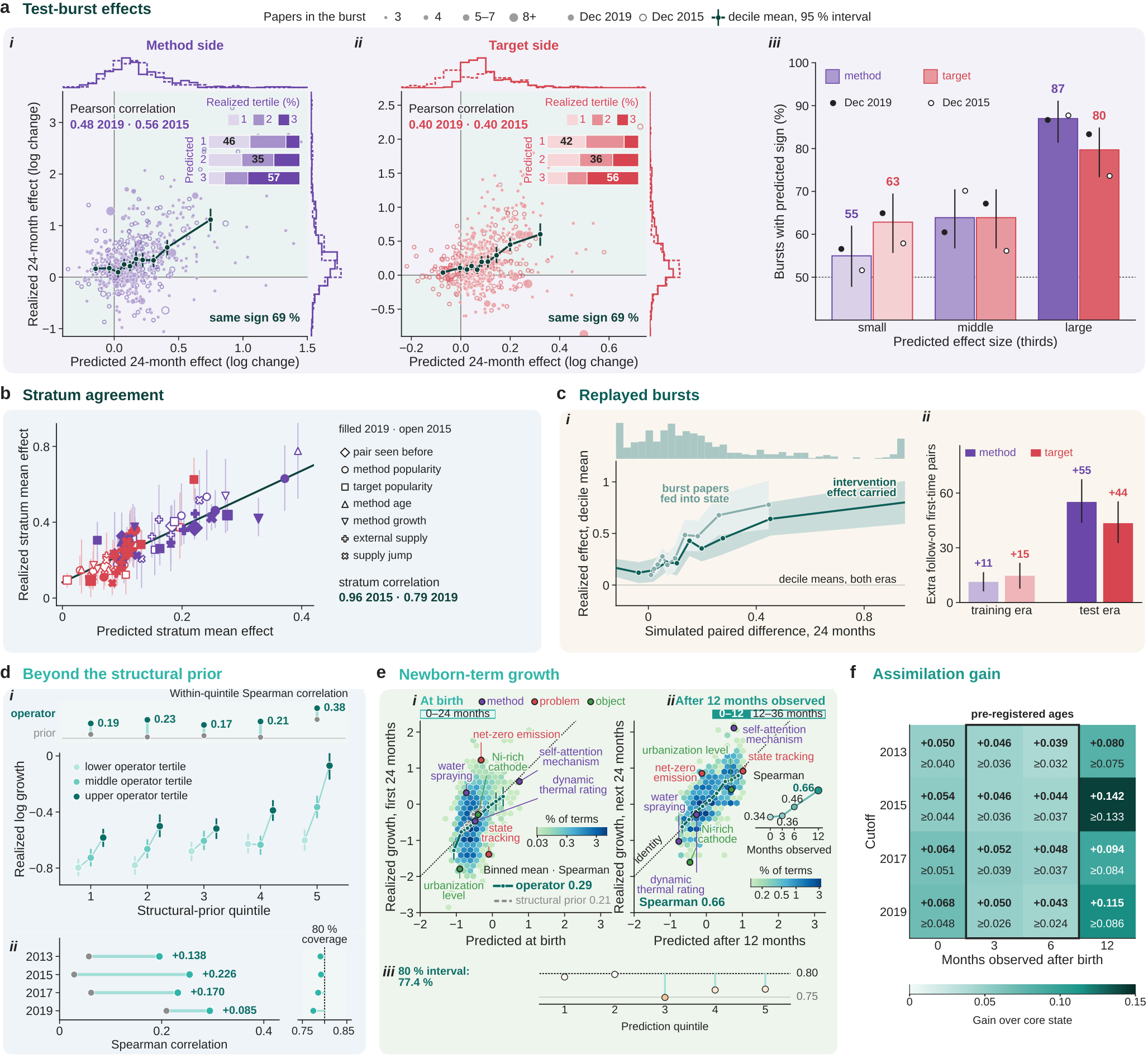}{Operators on held-out natural experiments (power engineering).}{fig:operators}{%
\textbf{a},~Intervention operator on 574 test bursts (filled, December 2019 cutoff; open, December 2015), predicted against realized 24-month effect relative to five matched controls. a(i), method side (Pearson 0.48 and 0.56). a(ii), target side (0.40 in 2019; 0.32--0.40 in 2015 across two seed sets). a(iii), share with the predicted sign by thirds of absolute predicted effect (all 574 bursts; method 55\,\% $\rightarrow$ 87\,\%, target 63\,\% $\rightarrow$ 80\,\%).
\textbf{b},~Stratum means, predicted against realized (correlation 0.96 in 2015, 0.79 in 2019; criterion $\ge 0.6$).
\textbf{c},~Transition operator replaying the bursts. c(i), realized effect by decile of the simulated paired difference for two paths. c(ii), follow-on first-time pairs per burst, training and test eras.
\textbf{d},~Entity-insertion operator, 3,472 terms first seen after December 2019. d(i), realized growth by structural-prior quintile and operator tertile. d(ii), Spearman correlation at four cutoffs and interval coverage.
\textbf{e},~Newborn forecasts at birth (e(i), entity insertion; Spearman 0.29 against 0.21 for the prior) and after 12 months (e(ii), observation assimilation; Spearman 0.66, $n = 857$). e(iii), coverage by quintile.
\textbf{f},~Observation-assimilation gain over the growth head of the real core state by age and cutoff.}

The transition operator was tested by replaying the test bursts. Two ways of propagating a burst (paths; \cref{fig:operators}c(i)), feeding its papers into the state and carrying the intervention effect forward, ranked the realized effects with Spearman correlations of 0.33--0.38 and intervals above zero, as did the direct growth head. The method-side ranking criterion held at every evaluable horizon from 12 to 60 months, and the follow-on first-time pairs per burst (+11 method-side and +15 target-side pairs in the training era, +55 and +44 in the test era) met the pre-registered criterion for a growing graph (\cref{fig:operators}c(ii)).

Because a new term has no history of its own, the entity-insertion operator was compared with a structural prior taken from the link cores. It ranked 24-month growth better than the prior at all four cutoffs, with gains in Spearman correlation of +0.085 to +0.226 (\cref{fig:operators}d,e). The operator's 80\,\% conformal intervals\cite{romano2019cqr} covered 0.774--0.792 of outcomes.

Once the first papers on a term have appeared, the observation-assimilation operator is applied instead. Three and six months after birth, it improved at all four cutoffs on the growth head applied to the term's real core state (\cref{fig:operators}f). After one year, its 24-month forecast reached a Spearman correlation of 0.66 (\cref{fig:operators}e(ii)). Its 80\,\% intervals covered 0.77--0.82 of outcomes at every cutoff. For single papers, the event-readout operator predicted 36-month citations, reuse and disruption better than LightGBM with text features in every run, and the operator results held under new seeds.

\subsection*{Natural-language what-if questions}\label{sec:res-whatif}\arxivtoc{subsection}{Natural-language what-if questions}

Each of the 17 what-if types in the registry, and the observation type, is labeled with the extent to which natural experiments identify its effect. For the demonstration observation question, the executor returned 24-month ratios to the baseline of 0.62--1.54 for three grounded actions, and the judge-model audits under three sampling seeds supported 54 of 56 claims. For the release of a dataset, code or tool, the model estimates the uptake of the concept that the release serves at +0.03 log, an effect confirmed on test events at both validation cutoffs. The model estimates that retracted papers receive in-field citations 0.25 log below those of matched papers. Both effects are partly identified.

Questions are compiled by the fine-tuned decoder at two levels: a question matched to a registered type is executed by the module of that type (level 1), and any other question is compiled into an action program over six state primitives, each executed by an existing operator (level 2; \cref{fig:whatif}a). On 449 new questions, 136 of them outside the registry, the decoder routed 0.878 of the questions exactly, against 0.768 for a general-purpose language model given the same registry (difference +0.109 \ci{+0.069}{+0.149}), and per-type routing thresholds raised exact routing on 187 further questions (difference +0.021 \ci{+0.005}{+0.043}; \cref{fig:whatif}b). The independent judge model passed 0.71 of 1,550 programs in 31 categories on all five of its checks (\cref{fig:whatif}c). Where historical events exist, the realized effects of programs were reproduced for adoption targets, whose precision exceeded that of a ranking by target popularity by +0.058 \ci{+0.017}{+0.096} over 117 test-era waves, and, in level, for dataset or tool releases (bias $-0.066$ \ci{$-0.171$}{+0.032}; \cref{fig:whatif}d).

\arxivfig{figure}{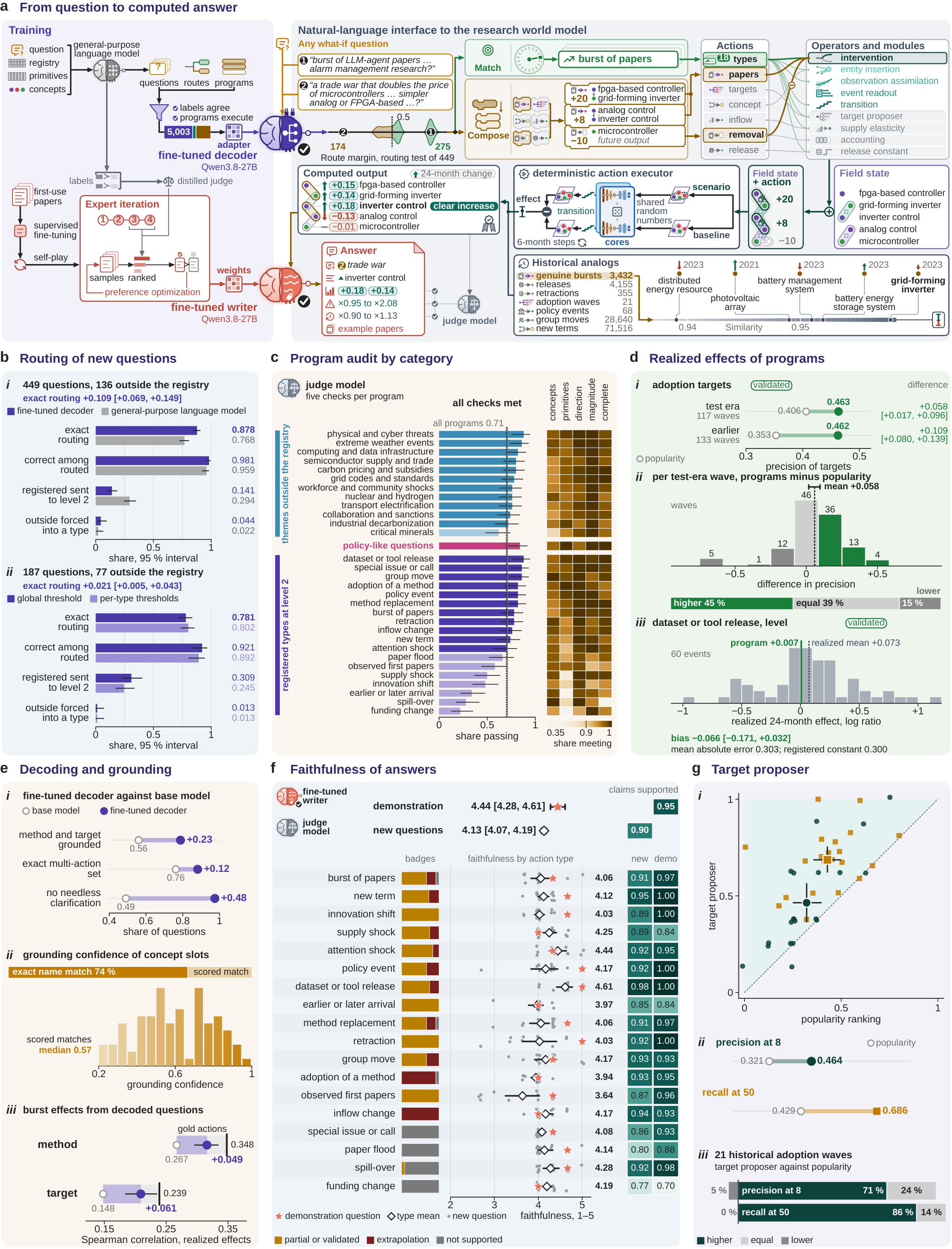}{From a what-if question to a faithful answer.}{fig:whatif}{%
\textbf{a},~From question to computed answer (schematic with system outputs). Left, training of the fine-tuned decoder (adapter on Qwen3.8-27B; 5,003 rows written and labeled by a general-purpose language model; routing threshold 0.5, $-0.25$ for six action types) and of the grounded answer generator (fine-tuned writer; supervised fine-tuning, self-play and four rounds of expert iteration, the last two ranked by judges distilled from labels of the independent judge model). Right, route margins of the 449 routing-test questions (at the threshold 0.5, the decoder routes 275 to level 1 and 174 to level 2). Example 1 is matched to a registered type, and example 2 is composed from state primitives, its removal of future output executed by the intervention operator as a burst with the opposite sign. The deterministic action executor runs scenario and baseline through the trained cores and the transition operator in 6-month steps with shared random numbers. Historical analogs, the five most similar past genuine bursts for example 2. The judge model checks each claim against the computed output.
\textbf{b},~Routing of new questions (95\,\% Wilson intervals). b(i), 449 questions, fine-tuned decoder against the general-purpose language model (exact routing 0.878 against 0.768, difference +0.109 \ci{+0.069}{+0.149}). b(ii), 187 further questions, global against per-type thresholds (exact routing +0.021 \ci{+0.005}{+0.043}).
\textbf{c},~Level-2 programs passing all five checks of the judge model, 50 programs per category (95\,\% Wilson intervals; dotted, 0.70; gray, all 1,550 programs, 0.71). Heat map, share of programs meeting each check.
\textbf{d},~Realized effects of the two validated level-2 program types on historical events, with the event type hidden from the decoder. d(i), precision of the adoption targets against a ranking by target popularity (117 test-era and 133 earlier waves). d(ii), per-wave difference in target precision, programs minus popularity, over the 117 test-era waves (mean +0.058 \ci{+0.017}{+0.096}; share of waves higher, equal and lower). d(iii), realized 24-month effects of 60 dataset or tool releases against the constant release effect of the programs (+0.007; bias $-0.066$ \ci{$-0.171$}{+0.032}).
\textbf{e},~Decoding and grounding of registered actions against the base model. e(i), both slots grounded correctly (3,868 questions; 0.560 $\rightarrow$ 0.788), exact multi-action set (210; 0.762 $\rightarrow$ 0.881, adapter run used in the system) and no needless clarification (157; 0.490 $\rightarrow$ 0.975). e(ii), grounding confidence of 365 filled concept slots. e(iii), Spearman correlation of burst effects predicted from decoded questions with realized effects (bars, gain; whisker, 95\,\% interval of the gain; black line, gold actions).
\textbf{f},~Faithfulness of answers audited by the judge model, three sampling seeds. Top, the 18 demonstration questions (4.44 \ci{4.28}{4.61}) and 215 new questions (4.13 \ci{4.07}{4.19}). Rows, the 18 action types: badge mix of the new answers, per-question means (dots), type mean with 95\,\% bootstrap interval (diamond), demonstration question (star) and claims supported.
\textbf{g},~Target proposer on 21 historical adoption waves. g(i), per wave and mean with 95\,\% interval (circles, precision at 8; squares, recall at 50). g(ii), precision at 8 (0.464 against 0.321 for popularity) and recall at 50 (0.686 against 0.429). g(iii), share of waves with higher, equal or lower values than popularity (precision at 8: 15, 5 and 1).}

For registered actions, the fine-tuned decoder uses two low-rank adapters\cite{hu2022lora,dettmers2023qlora} on its base language model, one for actions and one for clarification requests. In end-to-end tests, the fine-tuned decoder grounded both method and target to the correct concept in 0.788 of questions, against 0.560 for the base model without adapters, and decoded more multi-action questions exactly (\cref{fig:whatif}e). On a new set whose labels were verified by the independent judge model, the decoder asked for clarification on as many ambiguous questions as the base model but on only 4 of 157 clear ones, where the base model asked 80 times. Burst effects predicted from its decoded questions also correlated more strongly with realized effects, consistent with its better grounding of method and target (\cref{fig:whatif}e(iii)).

On 215 new questions covering every action type, the independent judge model compared each answer with the executor output it was written from. Faithfulness was 4.13 \ci{4.07}{4.19} of 5 and 90\,\% of claims were supported (\cref{fig:whatif}f), and faithfulness was 4.10--4.17 in every identification class. The 18 demonstration answers, one per what-if type plus observation, scored 4.44 (\cref{fig:whatif}f).

For adoption questions, a target proposer, learned from link-core scores with a correction for target popularity, selects the targets of the method, and the intervention operator computes the effect on each method--target pair. The executor reports method-side quantities and target effects as separately labeled entries. The grounded answer generator writes each answer from these entries and states extrapolated effects by their direction. Answers to 62 adoption and group-move questions (one sampling seed) scored a faithfulness of 4.13, with 93\,\% of claims supported and an over-claiming rate of 0.048. Answers to 23 questions whose actions are all extrapolations scored 4.12, with an over-claiming rate of 0.10.

Questions about the adoption of a method are routed to this target proposer. Over 21 historical adoption waves, its precision at 8, the share of its top 8 targets that the method reached, exceeded that of a ranking by target popularity by +0.143 \ci{+0.083}{+0.202} (\cref{fig:whatif}g).

\subsection*{Field-scale scenarios}\label{sec:res-sims}\arxivtoc{subsection}{Field-scale scenarios}

Each scenario was run from the validation origins December 2015 and December 2019 as 12 paired trajectories under several mechanism families, with full dose curves (\cref{fig:simulations}a,b). Of 25 candidate directions, 19 met the robustness criterion (\cref{fig:simulations}a,c), and a flood of machine-generated papers changed five-year field entropy by more than run-to-run noise.

\arxivfig{figure}{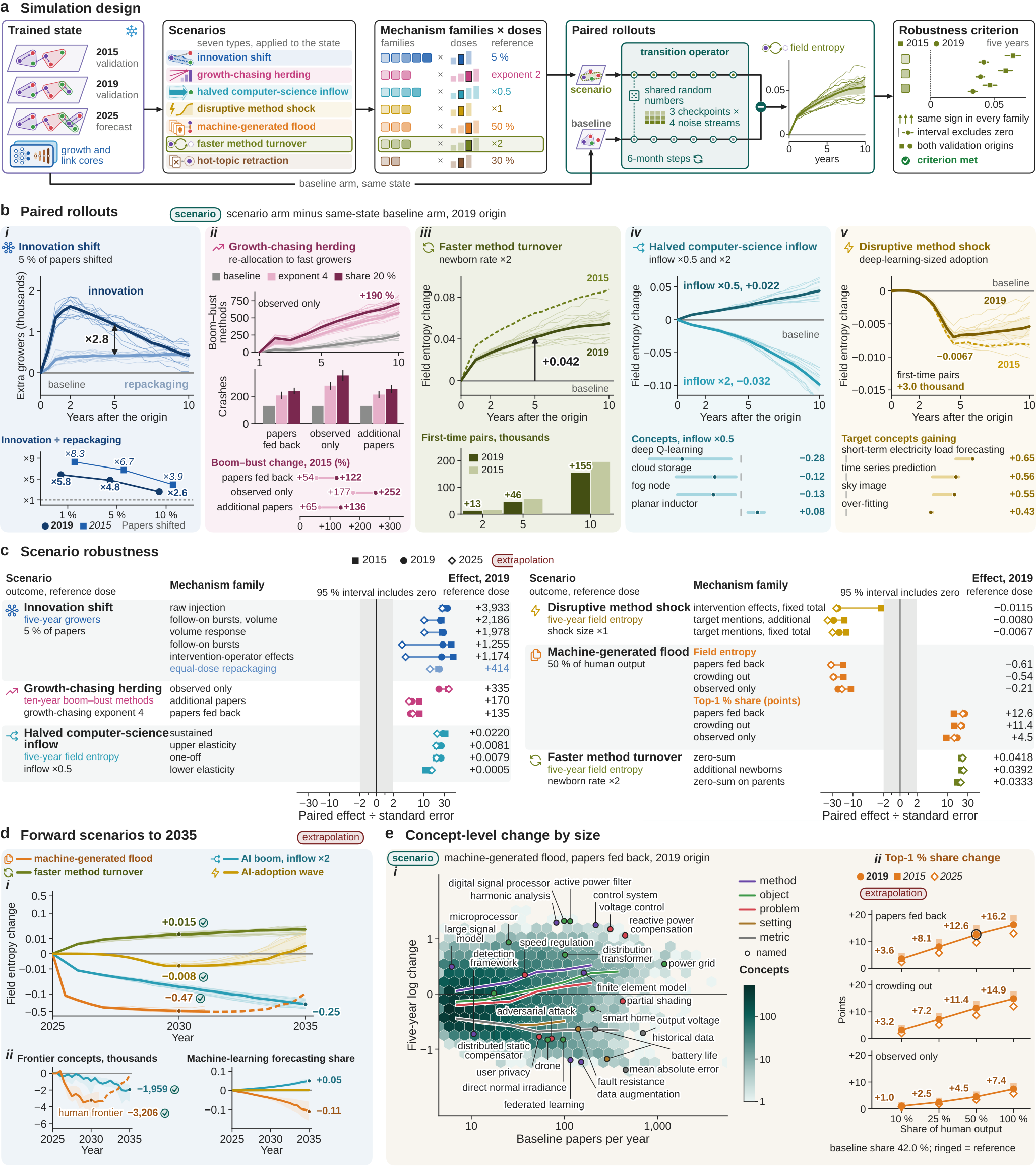}{Field-scale scenarios with the transition operator.}{fig:simulations}{%
\textbf{a},~Simulation design. The trained state (typed hypergraph field state with the trained growth and link cores) at the validation origins December 2015 and December 2019 and at the forecast origin December 2025 is perturbed by seven scenario types, each run under several mechanism families (tiles) and a dose ladder (bars; outlined, reference dose). The scenario arm and a baseline arm from the same state are rolled forward by the transition operator in 6-month steps with shared random numbers, giving 12 paired trajectories (3 checkpoints $\times$ 4 noise streams). Their scenario-minus-baseline differences are shown for faster method turnover (field entropy, 2019 origin; thin lines colored by checkpoint, mean and 95\,\% interval of the mean). A direction meets the robustness criterion only when every mechanism family gives the same sign with a 95\,\% interval excluding zero at both validation origins (example: five-year field-entropy effect of faster method turnover for its three families; square, 2015; circle, 2019).
\textbf{b},~Paired rollouts at the 2019 origin (thin, pairs; thick, mean). b(i), innovation shift (5\,\% of papers to first-time pairs), extra five-year growers against equal-dose repackaging ($\times$2.8); below, innovation $\div$ repackaging (median over families) against the share shifted at both origins. b(ii), growth-chasing herding, boom--bust methods (+190\,\% at ten years with a fifth of method mentions re-allocated; family without feedback), with ten-year crashes per family and the 2015 replication below. b(iii)--b(v), field-entropy change under faster method turnover, halved computer-science inflow (and its mirror $\times$2) and a disruptive method shock (dashed, 2015 mean), with cumulative first-time pairs (b(iii)) and named concepts with the largest changes (b(iv), b(v)) below. Every plotted direction meets the robustness criterion of a.
\textbf{c},~Paired effect over its standard error at the reference dose, per family and origin (herding at exponent 4; gray, interval includes zero).
\textbf{d},~Forward scenarios from December 2025 to 2035 (paired differences only). d(i), field-entropy change (AI boom, computer-science inflow doubled; AI-adoption wave, deep-learning-sized adoption path). d(ii), frontier concepts and the machine-learning-forecasting share.
\textbf{e},~Machine-generated flood. e(i), five-year change of human papers per concept (papers fed back, 2019 origin). e(ii), change in the top-1\,\% concept share against flood size per mechanism family at three origins (ringed, reference dose).}

Moving papers to first-time method--target pairs produced more five-year growers than repackaging the same number of papers at both validation origins, 2.8 times as many at 2019 (\cref{fig:simulations}b(i)). When a fifth of method mentions was re-allocated toward growing methods, boom--bust methods became 190\,\% more numerous at ten years (2019 origin, mechanism family without feedback; \cref{fig:simulations}b(ii)). Faster method turnover and a halved computer-science inflow raised field entropy, a disruptive method shock lowered it (\cref{fig:simulations}b(iii)--(v)), and the flood concentrated the observed literature. Retracting part of the papers of a hot topic reduced its five-year output roughly in proportion to the share removed.

From the forecast origin of December 2025, every direction that met the robustness criterion keeps its validation-origin sign, except innovation against repackaging in two of its five mechanism families. The other three still favor innovation. From this origin, the flood lowered field entropy by 0.47 at five years and a doubled computer-science inflow by 0.25 at ten, and faster turnover raised it (\cref{fig:simulations}d). Effects changed monotonically with dose, and in the families with feedback herding reversed sign at growth exponents of 0 and 1. A flood of half the human output raised the top-1\,\% concept share by 4.5--15.0 percentage points at the validation origins and by 3.3--9.7 from December 2025 (\cref{fig:simulations}e).

\subsection*{A prospective forecast}\label{sec:res-forecast}\arxivtoc{subsection}{A prospective forecast}

From December 2025, the model forecasts 20,740 active power-engineering concepts (\cref{fig:forecast}a,b), each with an 80\,\% interval and a separately reported era bias band. The largest gain in share of mentions to 2030 is forecast for the machine-learning-forecasting region (+5.2 percentage points; \cref{fig:forecast}a(i),c), led by retrieval-augmented generation, explainable AI, hybrid frameworks and computational overhead, whereas the power-electronics, power-quality and electrical-machines regions lose most. Large language models and quantum computing are forecast to reach 3.6 and 2.2 times their 2025 output by 2030 (\cref{fig:forecast}b(i)). In the transition-operator baseline, the top-1\,\% concept share reaches a simulated 48.6\,\% by 2035.

\arxivfig{figure}{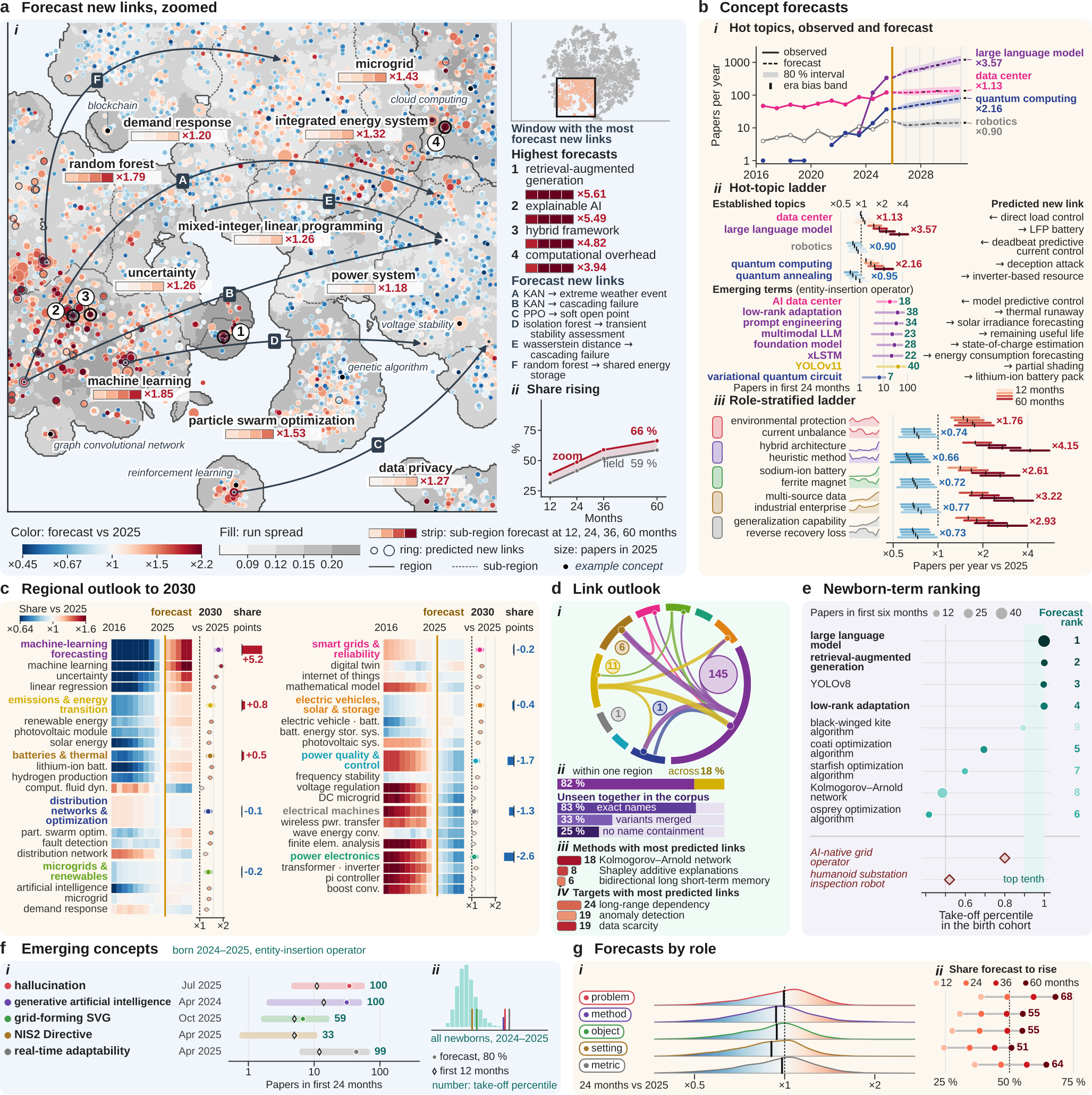}{Forecasts of power engineering.}{fig:forecast}{%
Forward outputs are from December 2025 (second origin March 2026). Rankings are those of the growth forecaster, and levels, intervals and era bias bands those of the five-run ensemble.
\textbf{a},~a(i), zoom of the field map (layout as \cref{fig:system}b) on the window with the most forecast new links (base map, current forecast). Color, 24-month forecast versus 2025. Size, papers in 2025. 1--4, the highest-ranked concepts with at least 20 papers in 2025. A--F, method$\rightarrow$target links forecast at December 2025 that no paper before 2026 connects (KAN, Kolmogorov--Arnold network; PPO, proximal policy optimization). Fill, run spread; strips, sub-region forecasts at 12--60 months; rings, concepts with predicted new links. a(ii), share forecast to rise.
\textbf{b},~b(i), hot topics, observed and forecast with 80\,\% interval and era bias band. b(ii), established topics and emerging terms first used in power and energy research in 2023 or later (entity-insertion operator), with each topic's top predicted new link at December 2025 (arrow, from method to target; new, target outside the 5\,\% of targets linked to the most methods, connected to the topic by no paper before 2026). LFP, lithium iron phosphate; LLM, large language model. b(iii), highest- and lowest-ranked concept per role.
\textbf{c},~Regional outlook to 2030: share of mentions relative to 2025, 2030 multiplier and share change by region and sub-region (abbreviations: batt., battery; comput., computational; conv., converter; dyn., dynamics; elem., element; optim., optimization; part., particle; pwr., power; stor., storage; sys., system).
\textbf{d},~The 200 likeliest unlinked pairs. d(i), regions connected. d(ii), share within one region (82\,\%) and share of the top 200 unseen together in the corpus under three nested criteria (exact names; variants merged, spelling and synonym variants; no name containment, also no paper with concepts whose names contain both). d(iii), d(iv), methods and targets with most links.
\textbf{e},~Newborn terms of 2024--2025 (corpus) whose first use in power and energy research is 2023 or later (literature check), ranked by the entity-insertion operator from March 2026: take-off percentile in the birth cohort (dotted, median) and rank of the 24-month growth forecast. Diamonds, two hypothetical terms.
\textbf{f},~Newborns of 2024--2025 (corpus) whose first use in power and energy research is 2023 or later (literature check). f(i), top newborn per role. f(ii), distribution.
\textbf{g},~g(i), 24-month forecasts by role. g(ii), share forecast to rise by horizon.}

The link core forecasts method--target links that no earlier paper had made, among them Kolmogorov--Arnold networks for extreme weather events and cascading failures, proximal policy optimization for soft open points and isolation forests for transient stability assessment (\cref{fig:forecast}a). Of the 200 highest-ranked pairs, 83\,\% have never been mentioned together in the corpus under their exact concept names, 33\,\% under any spelling or synonym variant and 25\,\% when name containment is also counted, and 82\,\% join concepts of one map region (\cref{fig:forecast}d). From the March 2026 state, the entity-insertion operator ranks large language model, retrieval-augmented generation, YOLOv8 and low-rank adaptation first among recent newborn terms (\cref{fig:forecast}e). Among the newborn terms of 2024 and 2025, hallucination and generative artificial intelligence lead the problem and method roles (\cref{fig:forecast}f). At 24 months, the median forecast of every role is close to no change, at 0.90--0.99 times the 2025 level (\cref{fig:forecast}g(i)), and the share of concepts forecast to grow rises with the horizon, to 51--68\,\% at 60 months (\cref{fig:forecast}g(ii)).

\section*{Discussion}\label{sec:discussion}\arxivtoc{section}{Discussion}

Research world models, as introduced here, draw forecasts and what-if answers from one transition of a research field, learned from its publication record. Earlier forecasts of research fields predicted new links in concept networks built from the literature. Krenn et al.\cite{krenn2023forecasting} forecast which pairs of artificial-intelligence concepts would first appear together in papers, and Marwitz et al.\cite{marwitz2026predicting} used concept graphs extracted by language models to propose new research directions in materials science, which domain experts then assessed. Arman et al.\cite{arman2026dynscigraph} combined temporal graph embeddings and language models in a dynamic knowledge graph and evaluated the pipeline on simulated corpora. The first two studies join concepts that occur together in a paper, and none of the three studies estimated how a field would respond to an intervention. Our hypergraph types each concept by its role in the paper. A forecast therefore names which method will reach which problem, and these forecasts were scored against hard negatives and with spelling and synonym variants merged. The operators add what-if answers drawn from the same core states as the forecasts, so that both can be tested against one publication record. The training scheme and three of the operators were also validated in two further fields, semiconductors and materials science.

The baselines were matched to the training window of the forecaster, and two families of them, LightGBM and a multilayer perceptron, were tuned, because forecasting margins shrink as baselines become stronger. The publication record also has to be cleaned before it can support event studies. Most bursts detected before deduplication were re-indexed copies of one work, and event studies on the uncleaned record would largely measure indexing. Bibliographic databases that merge preprints, proceedings and journal versions are likely to need the same step.

An analysis of funding shows why a forecast alone cannot answer a what-if question. The apparent effect of a funder entering a topic was matched by its placebo, which indicates that funders enter topics that are already growing, and the registry returns no estimate for funding. To be answered with an identified effect, a type of action needs its own natural experiment and a statement of how far its effect is identified. Questions outside the registry are answered by programs over state primitives, and these answers carry the badge ``generated'' (Methods). Faithfulness was highest when the executor produced every number and the language models only translated questions and wrote answers from its output. The same division of labor may suit other systems that put a language interface on a quantitative model.

For users of the answers, the main limitation is that the operators rank effects well but underestimate their size. Effects and simulated scenarios are reported as rankings and directions. Concept intervals are calibrated only for the most active concepts. The faithfulness audits rest on an independent judge model, which may favor fluent text\cite{zheng2023judging,panickssery2024llm}. Two further judges, one of them from an unrelated model family, rated the answers higher but agreed only weakly with it on individual answers. The faithfulness scores therefore depend on a single judge model, whereas the forecasts were scored against observed data. Changes in inflow from neighboring fields are answered only as extrapolations, and answers to questions outside the registry are not validated against natural experiments. The full system exists only for power engineering. Links between concepts other than a method and its target are forecast only as first co-occurrence, without a relation type, and there the readout's margin over LightGBM on the same pair features is small. For new terms the gap is larger: the entity-insertion operator ranks the fast-growing terms of the AI era first but underestimates how large they become, and three pre-registered corrections did not change this. We attribute the gap to a take-off that lies outside the support of the training data. We will score the December 2025 forecast against the papers published in 2026 and 2027 as these data become available.

\section*{Methods}\label{sec:methods}\arxivtoc{section}{Methods}

\subsection*{Corpus and field state}\label{sec:m-corpus}\arxivtoc{subsection}{Corpus and field state}
Power-engineering works were harvested from 33 OpenAlex topics\cite{priem2022openalex}, with every article or preprint published from 2000 onwards. After cleaning and excluding seven records dated after 2026, 1,196,720 abstracts remain. Deduplication by canonical work reduces them to 1,142,498 papers, 1,108,269 of them published in 2000--2025. With the citation layer the corpus holds 1.55 million papers and 11.13 million within-corpus citation events. The reader, a Qwen2.5-7B\cite{qwen2024qwen25} student low-rank-adapted from a 27-billion-parameter teacher, extracts concepts from the title and abstract, each with one of five roles (problem, method, object, setting, metric), and the typed relations between them, using greedy decoding and up to 400 new tokens. Concepts with at least five papers form a vocabulary of 204,287. Deduplication removed 54,222 duplicate records, and year-only dates were treated as a completeness problem.

The state $S_t$ of the research world model (equation~(\ref{eq:transition})) comprises the deduplicated typed hypergraph at month $t$ and the concept states that the growth and link cores compute from it. The hypergraph holds role-specific concept counts, typed method$\rightarrow$problem and method$\rightarrow$object edges and relations, an external-supply layer (each term's use in 58.5 million papers of seven other OpenAlex fields) and a citation-attention layer (the citations received by each concept's papers). Concept text embeddings are time-respecting yearly 64-component principal-component projections of SPECTER2 vectors\cite{singh2023scirepeval}.

\subsection*{Growth and link cores}\label{sec:m-cores}\arxivtoc{subsection}{Growth and link cores}
Both cores read the 100,000 most active concepts in $K = 8$ graph snapshots spaced 6 months apart. Two message-passing layers over co-occurrence and typed-link neighbors\cite{hamilton2017inductive} feed a temporal transformer\cite{vaswani2017attention,rossi2020tgn} over node states. Training runs for 6 epochs, and each head is calibrated separately. The growth core has heads at $H = 12$, 24, 36 and 60 months. The link core adds a graph stream, joined to the temporal stream through a learned gate, concept text and a typed-link head that adds a linear floor on 60 pair features to a residual network on the two concept states. The cores are trained separately because joint training degraded both.

In the default scheme every head is trained only on anchors with $t + 60 \le c$. Horizon-specific training cutoffs extend the training anchors to $t + 12 \le c$ for cutoff $c$. The loss of each $H$-head is then masked to anchors with $t + H \le c$ (links: $t + 24 \le c$), and each head is calibrated on its last 12 observed anchors.

One readout on the trained states of each link-core run scores candidate pairs for all 15 combinations of the five concept roles, and the scores of the five runs are averaged. For method$\rightarrow$problem and method$\rightarrow$object links the score is the core's typed-link score plus $\alpha$ times a residual network (two layers of 256 units, last layer initialized at zero). The residual receives the products and absolute differences of the two concepts' graph-stream states, 62 pair-law features, the core's score and the logarithm of its rank among the candidates of the anchor, and SPECTER2 embeddings of the two concept names. It is trained with a softmax loss over the candidate list of each of 11 anchors from 84 to 24 months before the cutoff. Each list is a uniform sample of 200,000 pairs from the full candidate list, and positives therefore keep their natural rate. The weight $\alpha \in \{0, 0.25, 0.5, 0.75, 1\}$ is chosen at the anchor 24 months before the cutoff, with the residual refitted on earlier anchors. The other 13 role pairs, and the first co-occurrence of methods with problems or objects, are scored by additive pair heads on the same states.

\subsection*{Operators}\label{sec:m-operators}\arxivtoc{subsection}{Operators}
The operators apply the actions $a_t$ of equation~(\ref{eq:transition}) to the state or read their effects from it. Because a new term has no history, the entity-insertion operator, a five-seed bagged LightGBM, uses early activity, external supply, the references of the first papers and a structural prior: the growth head of the link cores applied after the term is attached to its three text-nearest parents of the same role. Conformalized quantile regression\cite{romano2019cqr} gives its 80\,\% intervals, and take-off is reported only as a cohort percentile. Once papers on the term exist, the observation-assimilation operator, also a five-seed LightGBM, reads its real core state at ages 0, 3, 6 and 12 months, early activity, supply and the out-of-fold entity-insertion prior. Single papers are routed to the event-readout operator, a text-aware network that predicts 36-month citations, reuse of first-use pairs and disruption\cite{funk2017dynamic,park2023disruptive}.

The intervention operator averages difference-in-differences scores from LightGBM, a perceptron and counterfactual regression\cite{shalit2017estimating}, all trained on genuine bursts ($\ge 2$ distinct works in 3 months, 5 matched controls). Its level is a doubly robust\cite{robins1994estimation} estimate anchored to a fixed pool of 388 reference events (186 at the December 2015 origin). Its predictions are deterministic. Bursts are detected on the deduplicated record, and their outcomes, like the default-scheme core states that the operator reads, are derived from the graph before deduplication. The transition operator is a separately trained network with the core architecture (graph encoder and temporal transformer), applied autoregressively in negative-binomial 6-month steps with constant drift and shared random numbers. Its graph grows at a follow-on rate calibrated in the training era (11.3 method-side and 14.7 target-side pairs per genuine burst). No operator is gradient-coupled to the cores. The simulations use cores trained in the default scheme.

\subsection*{Interfaces}\label{sec:m-interfaces}\arxivtoc{subsection}{Interfaces}
The fine-tuned decoder is one base model, Qwen3.8-27B\cite{qwen2026qwen38card,qwen2025qwen3}, a 27-billion-parameter language model, with three low-rank adapters: actions, clarification, and routing and compiling (described below). For registered types it is served with vLLM\cite{kwon2023vllm} in FP8 and prompted with the schema, rules and six examples. In two passes, the actions adapter (QLoRA\cite{dettmers2023qlora} rank 16, trained on 3,084 rows that include multi-action questions) supplies the actions and the clarification adapter supplies the clarification note. A name is grounded by exact match or, failing that, to the five most similar concepts of the same role, with similarity weighted 0.6 on character-$n$-gram TF-IDF and 0.4 on SPECTER2 embeddings. A language-model variant judge maps spelling and synonym variants of new terms to existing concepts. The deterministic action executor runs scenario and baseline with shared random numbers, retrieves $k = 5$ historical analogs and attaches the badge of the action type. It describes changes in words ($|d| < 0.05$ ``roughly flat'', 0.05--0.18 ``slight'', $\ge 0.18$ ``clear'') and labels the era bias band explicitly as a bias. The grounded answer generator (fine-tuned writer) is the same base model fine-tuned by supervised fine-tuning, self-play and four rounds of expert iteration\cite{anthony2017thinking}, the last two ranked by judges distilled from the independent judge model. It writes six fixed sections conditioned only on the executor's result. The independent judge model (Claude Opus, Anthropic) scored faithfulness (1--5), claim support, over-claiming and section completeness against the executor output, seeing each answer in a separate call without the question or the arm label. It also audited the level-2 programs, and none of the reported effect sizes or forecasts was computed by it. Questions outside the registry are routed by the routing and compiling adapter of the fine-tuned decoder (rank 16) to programs over six state primitives (injected papers, removed papers, scaled inflow, an added concept, adoption targets from the target proposer and a constant release effect). The adapter was trained on 5,003 routes and programs written by a general-purpose language model. The operators execute these programs as paired scenarios under the badge ``generated''. An added concept is forecast by the entity-insertion operator, and answers carry the historical analogs of each injected pair, retrieved as for registered actions. On 449 new questions its routing was exact for 0.878, against 0.768 for a general-purpose language model given the same registry. Realized effects of the programs were validated for adoption targets and, in level, for releases. These tests are counted separately from the 42 tests of the evaluation protocol.

\subsection*{Evaluation protocol}\label{sec:m-protocol}\arxivtoc{subsection}{Evaluation protocol}
The cores were evaluated at walk-forward cutoffs from December 2013 to December 2021 in two-year steps. The operators were evaluated at December 2015 and 2019, with 2013 and 2017 as supplements. The evaluation cutoffs, the last month with complete data (March 2026) and the origin of the prospective forecast (December 2025) lie on one walk-forward time axis (\cref{fig:forecasting}a). ``Test'' denotes events after the cutoff, used neither for training nor for any choice of model or hyperparameters. The growth metric is the Spearman correlation over the 1,000 most active concepts. Restricting it to these concepts excludes the regression to the mean that dominates the growth rates of rarely mentioned concepts. For links the metric is typed average precision. Positives are first-time typed method$\rightarrow$problem or method$\rightarrow$object links formed within 24 months of an anchor, and negatives are sampled method--problem or method--object pairs not linked by then, either overall or restricted to hard negatives (pairs without a common neighbor at the anchor). For links among all role pairs, the metric is average precision over all new candidate pairs, the pairs among the 1,000 most active concepts of each role that no earlier paper had joined under any variant of their names, with a link counted at the first paper that mentions both concepts within 24 months. Labels are scored only if $t + H \le$ March 2026, because truncated windows inflate long-horizon advantages (by about 0.01--0.03). The baselines are default, horizon-matched and tuned LightGBM\cite{ke2017lightgbm} and a tuned perceptron on the same node features, 11 link-prediction baselines for links among all role pairs and Qwen3.8-27B\cite{qwen2026qwen38card,qwen2025qwen3} used zero-shot.

All training runs used deterministic kernels, at about 1--1.5\,\% extra run time, except three growth-core runs at each of the December 2015, 2019 and 2021 cutoffs. Single runs of the horizon-specific growth core have a standard deviation of 0.012--0.054 (median 0.023). Every comparison therefore bootstraps over training runs (re-ensembled) and anchors ($B = 2{,}000$--$4{,}000$). Non-inferiority bounds were $-0.005$ for switching operator inputs to horizon-specific core states and $-0.01$ for horizon-specific cutoffs at horizons not required to improve. Before every evaluation, the question, data, primary metric, decision criterion and expected power were written into a version-controlled, time-stamped pre-registration log. The reported claims rest on 42 pre-registered tests, of which 28 met their criterion, 4 met it in part and 6 failed, and the other 4 were planned analyses without a criterion. In the forecast from December 2025, concepts are ranked by the growth forecaster (a rank blend of a ten-run horizon-specific ensemble with the tuned perceptron), and levels, intervals and the era bias band (the range of calibrated bias over the validation cutoffs, not a forecast range) are taken from a five-run ensemble of cores trained to March 2026. New pairs are pairs whose two concepts no earlier paper mentions together under any variant of their names and for which no earlier paper contains concepts whose names include both. Pairs and newborn terms named in the figures and text were also checked against the literature dated before 2026 (Semantic Scholar, OpenAlex and a web search, judged by a language model on the retrieved records): no earlier paper connects a named pair, and each named newborn term was first used in power and energy research in 2023 or later.

\section*{Code availability}\label{sec:m-code-availability}\arxivtoc{section}{Code availability}
The complete source code for this work, including the corpus construction and concept-resolution pipeline, the world-model cores and operators, the fine-tuned decoder and executor, the evaluation and simulation scripts, and the scripts that generate the figures and tables, is available from the corresponding author upon reasonable request.

\section*{Acknowledgments}\label{sec:acknowledgements}\arxivtoc{section}{Acknowledgments}
The authors thank Oklahoma State University for providing the computational resources used in this work.

\section*{Author contributions}\label{sec:contributions}\arxivtoc{section}{Author contributions}
F.S. conceived the idea, designed the study, developed the methods, built the data pipeline and the models, performed the experiments and analyses, and wrote the manuscript. Y.Z. and L.L. supervised the work and reviewed the manuscript.

\section*{Competing interests}\label{sec:competing}\arxivtoc{section}{Competing interests}
The authors declare no competing interests.

\section*{Additional information}\label{sec:additional}\arxivtoc{section}{Additional information}
Correspondence and requests for materials should be addressed to Ying Zhang (\href{mailto:y.zhang@okstate.edu}{y.zhang@okstate.edu}).

\section*{Supplementary Information}\label{sec:si-release}\arxivtoc{section}{Supplementary Information}
The Supplementary Information will be released upon publication.

\clearpage
\bibliographystyle{unsrtnat}
\bibliography{refs}

\end{document}